\documentclass[letterpaper,times]{IONconf}
\usepackage[linesnumbered,ruled,vlined]{algorithm2e}
\usepackage{float}
\usepackage{subcaption}
\usepackage{graphicx}
\usepackage{amsmath}
\usepackage{url}
\usepackage{natbib}
\usepackage[hidelinks]{hyperref}
\title{Grid2Guide: A* Enabled Small Language Model
for Indoor Navigation}

\author{
    First~Author, Md.~Wasiul~Haque, \textit{University~of~Alabama}% <- this '%' removes a trailing whitespace
    \vspace{1mm} \\%
    Sagar~Dasgupta, \textit{University~of~Alabama}%
    \vspace{1mm} \\%
    Mizanur~Rahman, \textit{University~of~Alabama}
    }

\begin{document}

\maketitle

% biography section. The * indicates a section excluded from numbering.
\section*{biography}

% Biographies are defined as follows:
% \biography{Author name}{author biography text}

\biography{Md. Wasiul Haque}{received the B.Sc. degree in software engineering from the Islamic University of Technology, Gazipur, Bangladesh, in 2023. He is currently pursuing the Ph.D. degree in transportation systems engineering from The University of Alabama, Tuscaloosa, AL, USA. His research focuses on transportation, machine learning, navigation, and autonomous vehicle software security.}

\biography{Sagar Dasgupta}{received the B.Tech. degree in mechanical engineering from Motilal Nehru National Institute of Technology Allahabad, Prayagraj, India, in 2015, the M.S.
degree in mechanical engineering from Clemson
University, Clemson, SC, USA, in 2020, and
the Ph.D. degree in civil engineering from The University of Alabama (UA), Tuscaloosa, AL, USA, in 2024. He is a Postdoctoral Fellow at The University of Alabama. His research interests include GNSSbased autonomous vehicle cybersecurity, cyber–physical system security, transportation digital twin, and intelligent transportation.}

\biography{Mizanur Rahman}{received the M.Sc. and Ph.D. degrees in civil engineering (transportation systems) from Clemson University,
Clemson, SC, USA, in 2013 and 2018, respectively. He is an Assistant Professor with the Department of Civil, Construction and Environmental Engineering, The University of Alabama (UA), Tuscaloosa, AL, USA. His research focuses on transportation cyber-physical systems/transportation digital twin, and corresponding cybersecurity aspects.}

% The Abstract. The * indicates a section excluded from numbering.
\section*{Abstract}
Reliable indoor navigation remains a significant challenge in complex environments, particularly where external positioning signals and dedicated infrastructure are unavailable. This research presents Grid2Guide, a hybrid navigation framework that combines the A* search algorithm with a Small Language Model (SLM) to generate clear, human-readable route instructions. The framework first constructs a binary occupancy matrix from a given indoor map. Using this matrix, the A* algorithm computes the optimal path between origin and destination, producing concise textual navigation steps. These steps are then transformed into natural language instructions by the SLM, enhancing interpretability for end users. Experimental evaluations across various indoor scenarios demonstrate the method’s effectiveness in producing accurate and timely navigation guidance. The results validate the proposed approach as a lightweight, infrastructure-free solution for real-time indoor navigation support.

\section{Introduction}
Navigating complex indoor environments can be a time-intensive and disorienting experience. Unlike outdoor spaces, where landmarks, visibility, and satellite-based tools support orientation, indoor facilities often confine movement to corridors, stairways, or elevators, restricting visibility and increasing reliance on signage and prior familiarity. Among these, airports stand out as some of the most complex public indoor spaces due to their vast physical footprint, multilevel architecture, and constantly changing operational states. In addition to finding gates, passengers must adapt to tightly scheduled layovers, re-routed security checkpoints, and shifting boarding procedures. For travelers with language barriers, cognitive limitations, or mobility impairments, even simple tasks such as locating restrooms or baggage claims can become daunting, especially when compounded by real-time disruptions like gate reassignments or delays \cite{bosch2017flying, guerreiro2019airport}. Shopping malls, on the other hand, present a different yet equally challenging navigation problem. These commercial hubs frequently consist of multiple floors, interconnected corridors, and a mix of permanent and temporary installations (i.e., seasonal stalls, kiosks), making the spatial layout complex and less predictable. Visitors often rely on static "you are here" maps, which may even be outdated and difficult to interpret, especially for elderly individuals or tourists \cite{belir2013accessibility, jeong2018sala}. Hospitals, meanwhile, represent a particularly high-stakes use case for indoor navigation due to the urgency, emotional stress, and life-critical nature of the visits. These facilities typically have a non-uniform layout shaped by decades of ad hoc expansions, creating labyrinthine interiors with complex structures, ambiguous signage, and inconsistent floor transitions. This is further complicated for those with cognitive, visual, or mobility impairments, who may need additional support to interpret spatial cues or signage \cite{iftikhar2021human, jamshidi2020wayfinding}. Several studies have shown that navigation issues are not limited to patients; hospital staff also experience disorientation in large healthcare facilities. For instance, nearly one-third of first-time staff members report confusion about the layout of complex hospital campuses, and the need to frequently provide directions can cost staff hundreds of hours annually \cite{uab_wayfinding, mapspeople_hospital}.

Outdoor navigation has been revolutionized by satellite-based technologies, such as Global Positioning System (GPS). These systems, however, are ineffective indoors because satellite signals experience severe attenuation, blockage, and multipath distortion caused by walls, ceilings, and other structural barriers. As a result, indoor navigation remains a largely unsolved problem, particularly in terms of finding a general, scalable, and user-friendly solution. Existing solutions often rely on infrastructure-intensive methods, such as Wi-Fi fingerprinting, Bluetooth beacons, Radio-Frequency IDentification (RFID), or custom sensor arrays. Other methods leverage on-device sensors through Simultaneous Localization and Mapping (SLAM), which fuses data from cameras, LiDAR, or inertial sensors to construct a map while estimating position. While SLAM avoids external infrastructure, it requires significant computational resources and can struggle in visually repetitive or dynamic environments. Platforms like IndoorAtlas and Quupa deliver highly accurate positioning by leveraging magnetic fields or Bluetooth Low Energy (BLE) tags; however, these require dedicated deployment and calibration for each building \cite{hurtuk2019indoor, quuppa2019}. While somewhat effective, such systems are expensive to maintain, lack flexibility, and are often impractical for wide-scale adaptation in public venues.

In response to the limitations of traditional indoor navigation solutions, researchers have increasingly turned their attention to utilizing floor plans, static architectural diagrams that are already available in most public and commercial buildings, as a scalable and cost-effective foundation for navigation solutions. These maps are typically used for safety signage and navigation, but may contain useful spatial information about the internal layout of the floors of a building. However, accurately interpreting these diagrams requires a level of spatial reasoning and contextual understanding, which is not easily handled in real-time by conventional rule-based systems. All the above needs and limitations underscore the need for navigation support that extends beyond conventional maps and static and dynamic signage, offering more intuitive and personalized guidance.

The emergence of Large Language Model (LLM) agents, such as ChatGPT, Claude, and Gemini, has offered a transformative opportunity in this space. These models are not only capable of interpreting natural language but also have evolved to process and reason over images, making them powerful, multi-modal general-purpose problem solvers. By utilizing their ability to generalize across domains, LLMs can potentially act as an interactive indoor navigation assistant. Instead of relying on expensive infrastructures such as beacons or RFID systems, users could simply upload a floor plan and ask the LLM for directions to a specific room or points of interest (POI). However, this solution also has several drawbacks. To process the floormap image directly using LLM is resource-intensive and time-consuming. Image processing is the main bottleneck of using an LLM with floor maps directly. Even though LLMs have vast generalization capabilities, processing a complex image and extracting meaningful information from it remains a resource-intensive task. Each query can require up to 4-5 minutes to be processed, which is not usable for real-time cases using smart hand-held devices. Even with the most advanced reasoning, the LLM available to date, the accuracy is not particularly sufficient \cite{coffrini2025methodllmenabledindoornavigation}. The high computational requirement issue is more acute because of the limited computing resources of smart hand-held devices.

To address these issues, we present a Small Language Model (SLM)-based approach that offloads spatial reasoning to a lightweight graph-search technique and utilizes SLM solely to generate human-readable instructions from terse navigation instructions. This approach involves creating an obstacle layout of the indoor map, and then a planning search algorithm can be utilized to find a suitable path from the origin to the destination. The contributions of the paper can be summarized as follows:
\begin{enumerate}
    \item All spatial reasoning of an indoor map is offloaded to A* Search on a binarized occupancy grid, guaranteeing accurate optimal routes in each iteration with low computational resources. To the best of our knowledge, no prior work has combined classical path planning with LLM/SLM-based indoor navigation, making this a lightweight alternative to resource-intensive LLM reasoning.
    \item A mechanism for handling multi-layer buildings is incorporated by defining portal nodes (e.g., elevators, escalators, staircases) to enable seamless navigation across multiple floors. To the best of our knowledge, no prior indoor navigation framework has demonstrated lightweight multi-floor path planning combined with natural language instruction generation.
    \item Interactive and human-understandable navigation instruction is generated with the combination of A* Search and an SLM for a real-time application. Unlike prior systems that rely on static maps or rigid rule-based instructions, this is the first to leverage an SLM for natural language guidance in indoor navigation.
    \item Integration of A* Search with SLM enables efficient execution on standard commodity CPU/GPU hardware. This computationally optimized pipeline ensures broad deployability on handheld devices, making it well-suited for real-time personal navigation applications. Existing LLM-based approaches require expensive infrastructure or high-performance computing; to the best of our knowledge, no work has demonstrated such deployability on mobile-class hardware.
\end{enumerate}

The remainder of this paper is organized as follows. Section 2 reviews the existing challenges and current solutions in indoor navigation, highlighting their feasibility and limitations. Section 3 introduces the proposed framework that integrates A* search with a Small Language Model (SLM) for indoor navigation, detailing the underlying components and design. Section 4 describes the experimental setup used to evaluate the proposed pipeline, including test environments, runtime performance, and instruction accuracy. Section 5 discusses the evaluation results, comparing the approach against baseline methods. Finally, Section 6 concludes the paper by summarizing the findings and outlining potential directions for future research.

\section{Literature Review}
Several studies have examined the applicability of generic Location-Based Services (LBS) platforms, such as IndoorAtlas, Anyplace, Quuppa, and Google Maps for indoor navigation in enclosed environments \cite{ramani2014indoor, van2016real, wichmann2024determining}. While these platforms demonstrate usability within designated areas, they require customized deployment for each specific location \cite{8911759}. This localized setup process is often time-consuming and resource-intensive, limiting scalability across diverse environments. In contrast, many indoor spaces already possess detailed floor plans, offering a promising opportunity to develop more generalized and scalable navigation solutions that leverage existing infrastructure. Such solutions will eradicate the problems caused by the lack of signals in closed environments.

The current advancements in artificial intelligence (AI) have been greatly reflected in the usage of generative AI tools (e.g., ChatGPT). With great general problem-solving and image processing capabilities, LLMs are utilized to provide generalized solutions in different fields. Leveraging LLM for providing proper navigation instructions in indoor setups is still not widely explored by researchers. One recent study was conducted and observed by Coffrini et al. \cite{coffrini2025methodllmenabledindoornavigation} where the authors have taken 2-directional maps of various indoor locations (i.e., shopping malls and airports), preprocessed the maps, and utilized the LLM to generate specific navigation instructions between two different points. Their experimentation has uncovered some weaknesses in the LLM's ability to predict accurate navigation instructions. The model often provided unnecessary information and instructed a path that is not physically present, even utilizing a refined system prompt. LLMs can solve many complex tasks, but often are seen to stumble in terms of complex reasoning. Such experimentations can be categorized under 'Few-Shot Prompting' \cite{sahoo2024systematic}. The system prompt acts as a high-level directive that defines the model’s role, behavior, and expected response style, thereby shaping the interpretation of few-shot examples. Studies have shown that well-crafted system prompts significantly improve the consistency and contextual accuracy of the model's outputs \cite{reynolds2021prompt, ouyang2022training, wei2022chain}. However, this approach is not usable in real-world scenarios due to the slow image processing capabilities of the LLMs. Utilizing the advanced reasoning capability, each prompt takes 4-5 minutes to be processed, creating a significant delay and hindering real-time applicability in hand-held devices \cite{coffrini2025methodllmenabledindoornavigation}. The computationally expensive method also introduces the requirement for costly, computationally resource-intensive infrastructure.

Finding an optimized route in an indoor environment can be categorized as a classic path planning from. Efficient graph searching algorithms have been proposed in path planning scenarios in various aspects \cite{tang2021geometric}. In such approaches, an alternative graph-like map of the environment is created to invoke existing graph searching algorithms. Various methods have been proposed to model the environment, such as Voronoi diagrams \cite{candeloro2017voronoi}, visibility graphs \cite{chou2019optimal}, and grid structures \cite{fransen2020dynamic}. Conventional shapes are used to represent the environment in the grid structure method, and each grid is consistent, making it suitable for most indoor use cases. The path search algorithm is a method used to find the path between the origin and destination on a map. Based on the approaches, such algorithms can be divided into bionic algorithms, geometric algorithms, and interpolation-based algorithms. Bionic algorithms are random search algorithms simulating natural biological evolution. A group of algorithms, such as the particle swarm optimization (PSO) algorithm \cite{das2016hybridization}, the genetic algorithm (GA) \cite{santiago2017path}, and the ant colony (AC) algorithm \cite{xiong2019path}, have been applied to path search processes. Even though these are easy to implement, they often suffer from local optima, making them inappropriate for path planning in complex grid map environments. Interpolation-based algorithms utilize smooth curves, such as the B-spline curve \cite{park2007b} and Dubins curves \cite{dubins1957curves} to fit the entire path. The generated paths are smooth, but they do not fit well to irregular paths; hence, their application is mainly suitable for smoothing local or simple path problems. On the other hand, geometric search algorithms are widely used as path search algorithms. Most common examples include the Dijkstra algorithm \cite{wang2011application} and the A* algorithm \cite{guruji2016time}. The Dijkstra algorithm obtains the shortest path by traversing all the nodes in every iteration, which is computationally expensive. The A* algorithm is based on the Dijkstra algorithm that optimizes by adding a heuristic function, which was first proposed by Hart \cite{hart1968formal, howden1968sofa}. The A* algorithm is suitable for real-time usage due to its faster performance, and is therefore widely used in path planning.

In recent years, compact instruction-tuned transformer models have emerged as powerful tools capable of following complex prompts with high fidelity, while maintaining efficiency in terms of computation and memory. This makes them particularly suitable for real-time applications such as indoor navigation, where system responsiveness and resource constraints are critical. Instruction tuning has been shown to substantially improve the ability of language models to generalize across a variety of tasks. Wei et al. demonstrated that instruction‑tuned language models substantially outperform their untuned counterparts across a wide range of benchmarks \cite{wei2021finetuned}. This finding laid the groundwork for a new generation of efficient language models capable of strong zero-shot and few-shot performance. Google’s FLAN‑T5 family delivered state‑of‑the‑art performance with as few as 780 million parameters \cite{chung2024scaling}. This balance between performance and resource efficiency enables use in edge environments such as mobile devices or embedded navigation systems. More recently, Fu et al. evaluated FLAN‑T5 (780 M) on real‑world meeting summarization and found that it matches or exceeds zero‑shot GPT‑3.5, despite being an order of magnitude smaller \cite{fu2024tiny}. Other studies have echoed these observations. Frantar et al. introduced GPTQ, a quantization technique that reduces LLM memory footprint without substantial performance loss, enabling models like LLaMA and OPT to run efficiently on consumer hardware \cite{fransen2020dynamic}. Similarly, Dettmers et al. proposed LLM.int8() as a precision-optimized inference method, allowing larger models to execute on a single GPU with minimal degradation \cite{dettmers2022gpt3}. Moreover, the TinyLlama project has demonstrated that even smaller models, with under 1 billion parameters, can be fine-tuned for high performance in instruction-following tasks while maintaining fast inference speeds and small memory footprints \cite{zhang2024tinyllama}. These models enable seamless deployment on devices with limited compute power, including Raspberry Pi-level hardware or mid-range smartphones, making them ideal for navigation systems that aim to remain lightweight, offline-capable, and privacy-preserving. These indicate that a small, on-device SLM can be utilized to reliably convert the terse text outputs from the A* algorithm to human-understandable navigation instructions, without using any specialized, expensive infrastructure.

\section{Method}
\begin{figure}[!ht]
  \centering
  \includegraphics[width=\linewidth]{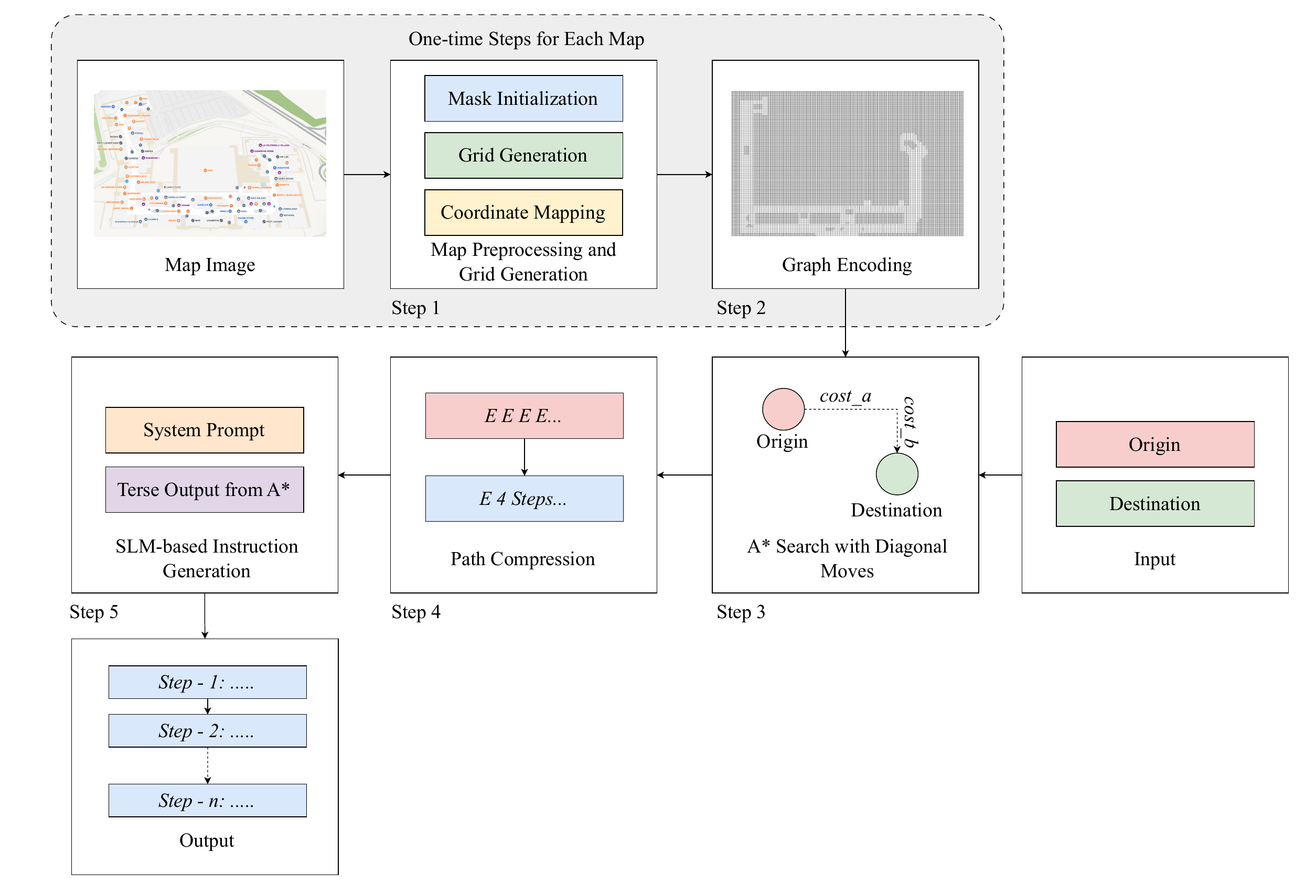}
  \caption{A* enabled SLM for indoor navigation.}
  \label{fig:pipeline}
\end{figure}

To enable computationally efficient and real-time indoor navigation, we present a five-stage pipeline that transforms a static floorplan image into fluent, human-readable directional instructions. From the user’s perspective, the system requires only two inputs: the origin and the destination on the map. As presented in Figure \ref{fig:pipeline}, our approach consists of the following stages: (i) Map Preprocessing and Grid Generation, (ii) Graph Encoding, (iii) A* Search with Diagonal Moves, (iv) Path Compression, and (v) SLM-Based Instruction Generation.

Finding a route between two points on a closed circuit with walkable free paths and obstacles is a path planning problem. The initial objective of such is to generate an obstacle-aware map from the provided floorplan image \cite{borges2019strategy}. To achieve this, we convert the image into a binary occupancy grid, classifying each pixel as either walkable or blocked using standard image thresholding and polygon-masking techniques. This preprocessing step yields a discrete 2D grid of free and occupied cells, effectively capturing the navigable space. Once the occupancy grid is generated, the A* search algorithm, augmented with 8-way (diagonal) movement, is employed to compute the shortest path between the specified start and goal coordinates. By offloading geometric reasoning to the A* algorithm, the pipeline eliminates the need for a machine learning model to directly process the image or infer spatial relationships. After obtaining the optimal cell-by-cell route, we apply a lightweight run-length encoding (RLE) scheme to compress repetitive movements and diagonal zig-zag patterns into a concise sequence of directional commands. This compression reduces redundancy and simplifies the input to the final stage.

The SLM is engaged in the last step. It receives the compressed route instructions and generates a fluent, numbered, and conversational walking guide. By restricting the SLM’s involvement to purely textual transformation, the system avoids costly image processing. It minimizes the risk of probabilistic navigation errors while still providing natural, intuitive instructions to the user.

\subsection{Map Preprocessing and Grid Generation}
\begin{figure}[!ht]
  \centering
  \includegraphics[width=.7\textwidth]{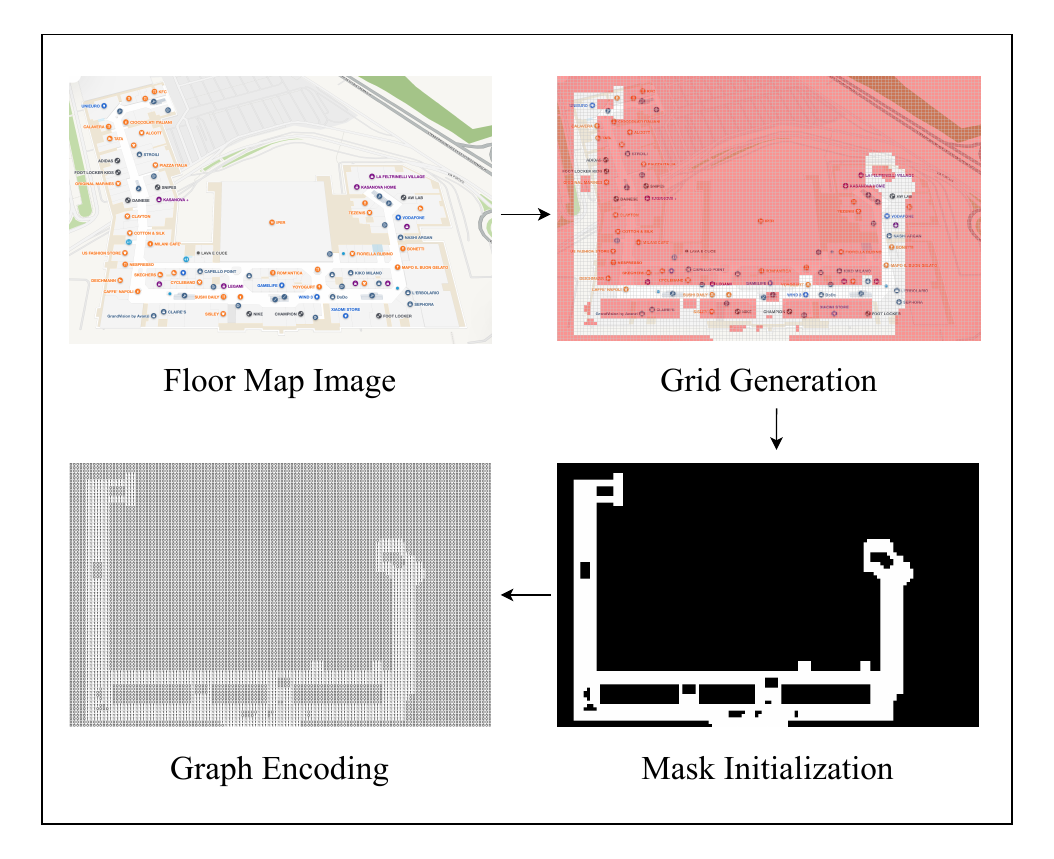}
  \caption{Map preprocessing prior to running the A* search.}
  \label{fig:map_processing}
\end{figure}

At the beginning, we take an existing floorplan image of any valid format (e.g., PNG, JPEG, TIFF). Any unnecessary parts (i.e., legends, annotations, etc.) that would not be helpful for navigation from the map image are cropped and removed. We utilize the map image to create a grayscale mask for binarization. This grayscale binary mask needs to be generated prior to running A* search and needs to be created only once for a specific map, and can be reused for every subsequent iteration of the A* search. Figure \ref{fig:map_processing} shows the entire steps of map image preprocessing for the floor plan of Orio Center, a shopping mall in Bergamo, Italy.

\begin{figure}[!ht]
  \centering
  \includegraphics[width=\textwidth]{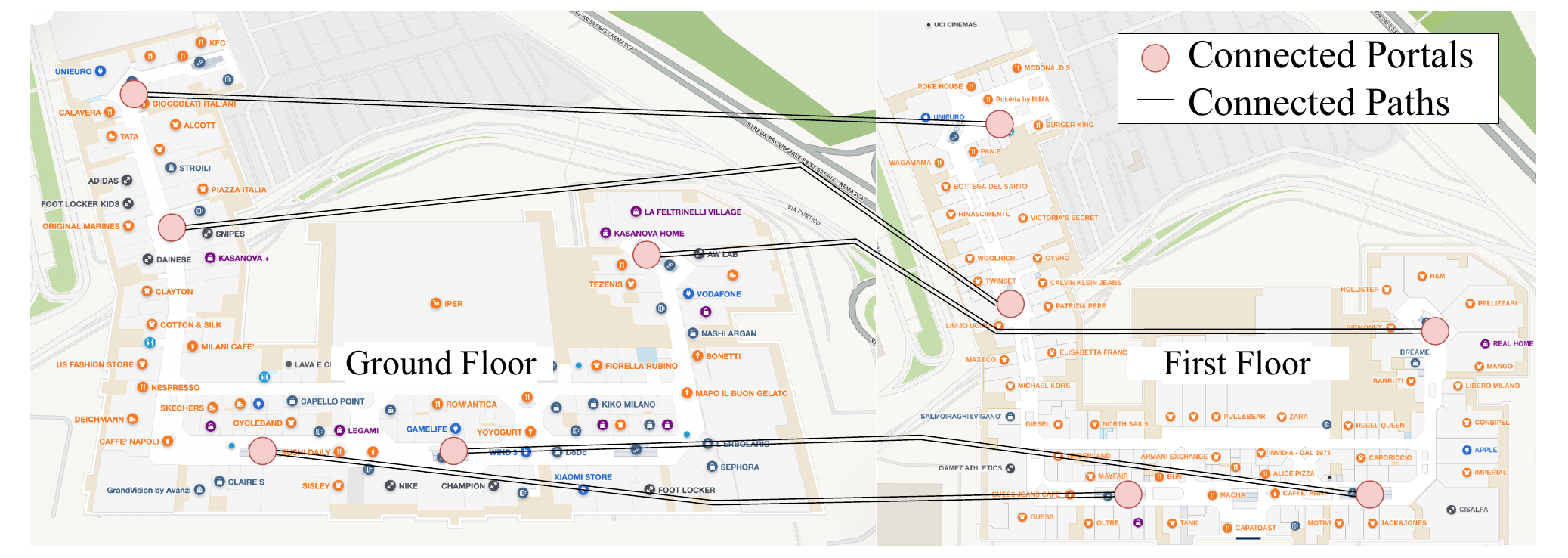}
  \caption{Portal nodes at the connected points between the ground floor and first floor of Orio Center, Bergamo, Italy.}
  \label{fig:multi_floor_node}
\end{figure}

The mask encodes walkable versus obstructed regions in the map. We load this mask and downsample it to a pre-determined $m \times n$ grid to produce a Boolean occupancy matrix $G$. For locations comprising multiple floors, a separate occupancy matrix is constructed for each level. To enable vertical traversal, portal nodes are manually defined at the locations of the connecting points (i.e., staircases, elevators, or escalators). Figure \ref{fig:multi_floor_node} shows an example case over two floors of Orio
Center, Bergamo, Italy. A portal is represented as a bidirectional link between two grid cells belonging to different floors, formally encoded as an edge in the navigation graph coordinates $(f_1, i_1, j_1)$ on floor $f_1$ to coordinates $(f_2,i_2,j_2)$ on floor $f_2$. The values of $m$ and $n$ for the Boolean occupancy matrix depend on the aspect ratios of the map.

  \begin{equation*}
    G \in \{0,1\}^{m \times n},
  \end{equation*}

where,

  \begin{equation*}
    G_{i,j} = \{^{1, \text{ if the pixels in cell $(i, j)$ are free,}}_{0, \text{ otherwise.}}
  \end{equation*}

\subsection{Graph Encoding}
In our implicit graph representation, every free cell $G_{i, j} = 1$ becomes a node. We connect it to its eight neighboring cells $(i+\Delta i, j + \Delta j)$ for $\Delta i, \Delta j \in \{-1, 0, 1\} \text{\textbackslash} \{0,0\}$. This 8-way connectivity is proposed to ensure smooth, diagonal shortcuts around sharp corners in the grid map, replicating human-expected movement. Table \ref{tab:movement_costs} enumerates all the 8-way connectivity and their assigned costs. To reflect movement cost, we assign each edge a weight:

  \begin{equation*}
    c(\Delta i, \Delta j) = \{^{1, \text{ } |\Delta i| + |\Delta j| = 1 \text{ (orthogonal move)}}_{\sqrt{2}, \text{ } |\Delta i| + |\Delta j| = 1 \text{ (diagonal move)}}
  \end{equation*}

\begin{figure}[!ht]
  \centering  \includegraphics[width=0.8\textwidth]{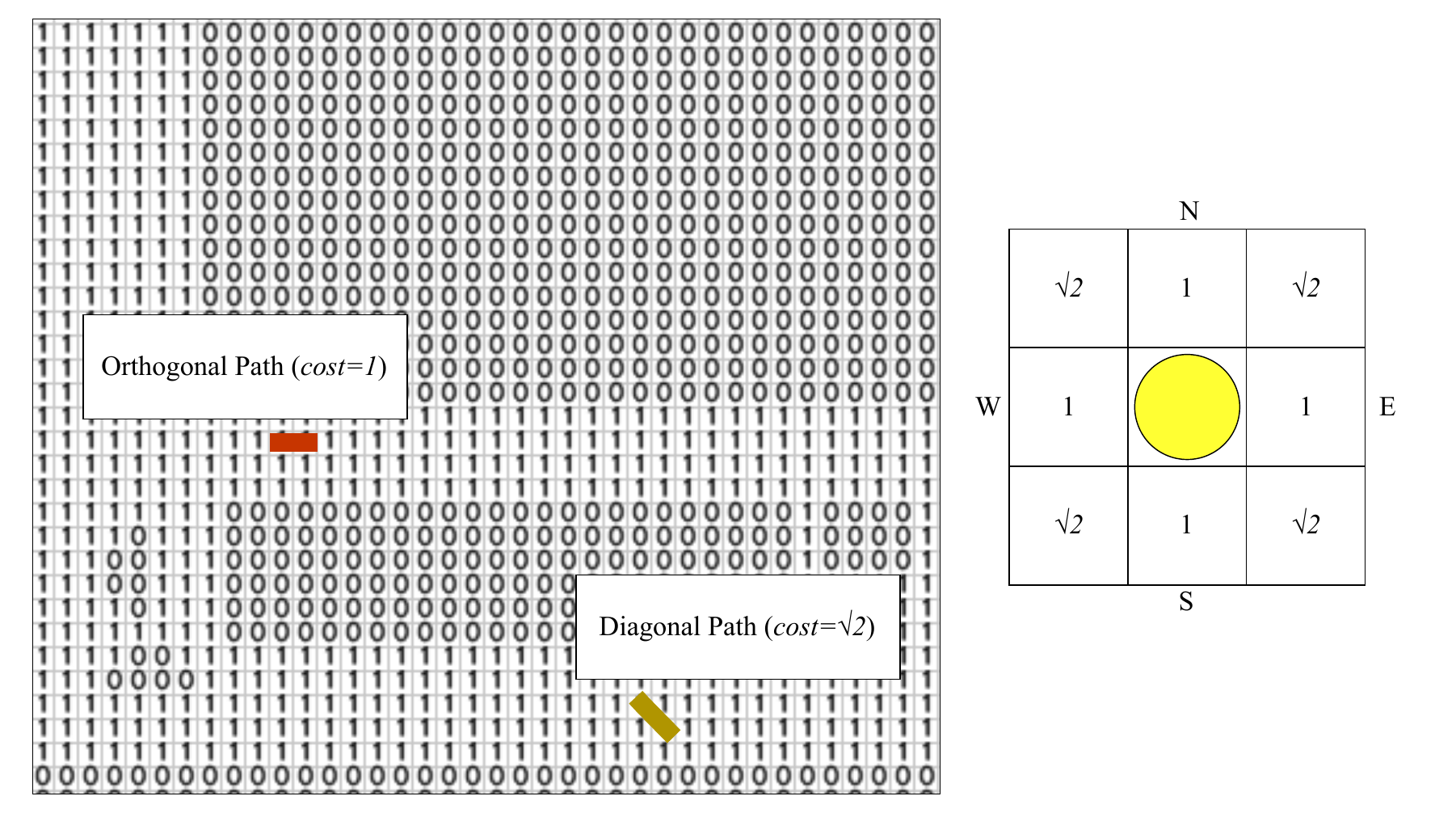}
  \caption{Different costs assigned for orthogonal and diagonal paths.}
  \label{fig:a_star_cost}
\end{figure}

Graphical representation of diagonal and orthogonal paths and their cost assignments between neighboring cells is presented in Figure \ref{fig:a_star_cost}.

\begin{table}[!ht]
  \centering
   \caption{Movement offsets and their associated costs in an eight‐connected grid for A* search.}
  \begin{tabular}{l l l}
    Direction & $(\Delta i,\Delta j)$ & Cost \\ 
    \hline
    N   & $(-1,0)$   & 1        \\ 
    S   & $(1,0)$    & 1        \\ 
    W   & $(0,-1)$   & 1        \\ 
    E   & $(0,1)$    & 1        \\ 
    NW  & $(-1,-1)$  & $\sqrt{2}$ \\ 
    NE  & $(-1,1)$   & $\sqrt{2}$ \\ 
    SW  & $(1,-1)$   & $\sqrt{2}$ \\ 
    SE  & $(1,1)$    & $\sqrt{2}$ \\ 
    \hline
  \end{tabular}
  \label{tab:movement_costs}
\end{table}

\subsection{A* Path Planning with Diagonal Moves}
A* is a classical best-first search algorithm that combines the exhaustive optimality of Dijkstra's method with heuristic guidance to dramatically reduce the number of nodes required to examine for planning an optimal path \cite{hart1968formal}. This search algorithm is both (i) complete: it will surely find a path if one exists, (ii) optimal: it will return the most cost-optimal, given that the heuristic is admissible and consistent. In our grid-based search formulation, A* search maintains three quantities for each free cell $(i, j)$:
\begin{itemize}
    \item Cost-to-Come, $g(i,j)$: The exact cost of the shortest known path from a cell to another. Whenever the A* considers stepping from a node $(i,j)$ to a neighbor $(i',j')$, it computes:
    
      \begin{equation*}
      g = g(i,j) + c((i,j) \rightarrow (i',j')),
      \end{equation*}

    where $c=1$ for orthogonal moves and $c=\sqrt{2}$ for diagonal moves. If $g$ improves upon the previously recorded cost at $(i',j')$, the algorithm updates $g(i',j')$ and records $(i,j)$ as its predecessor.

    \item Heuristic Estimate, $h(i,j)$: To search forward to the goal, avoiding blind exploration, the A* algorithm adds an estimate of the remaining cost from $(i,j)$ to the target cell $t = (i_t, j_t)$. We used the Chebyshev distance,
    
      \begin{equation*}
      h((i,j), (i_t,j_t)) = \text{max}\{|i-i_t|, |j-j_t|\},
      \end{equation*}

     which perfectly aligns with our mixed-cost model. Since this heuristic never overestimates the true minimal cost, it is both admissible and consistent.

     \item Total Score, $f(i,j)$: Finally, the A* algorithm assigns each node a priorty
     
      \begin{equation*}
      f(i,j) = a(i,j) + h(i,j),
      \end{equation*}

     and always expands the node with the smallest $f$ value next. The Cost-to-come, $g$, ensures that the already explored cost is exact, while the Heuristic Estimate, $h$, biases the search in the direction of the goal. This balance allows the search to converge on the optimal path quicker than uninformed search methods like breadth-first or Dijkstra's search, particularly on large or complicated grids.
\end{itemize}

By integrating these three variables in a simple min-heap, our implementation of the A* algorithm computes guaranteed optimal indoor routes. The formal sequences of the implementation are presented in Algorithm \ref{alg:astar}.

\begin{algorithm}[!ht]
\caption{A* Path Planning with 8‑Way Connectivity}
\label{alg:astar}
\KwIn{Grid $G$, start node $s$, target node $t$}
\KwOut{Optimal path from $s$ to $t$}

$N_8 \gets \{(-1,0,1),(1,0,1),(0,-1,1),(0,1,1),(-1,-1,\sqrt{2}),(-1,1,\sqrt{2}),(1,-1,\sqrt{2}),(1,1,\sqrt{2})\}$\;
$h(u,v) \gets \max(|u.i - v.i|,\ |u.j - v.j|)$ \tcp*[r]{Chebyshev Heuristic}

Initialize empty min-heap \textit{openSet}\;
Push $(h(s,t),\,0,\,s)$ into \textit{openSet}\;
$g[s] \gets 0$\;
$came\_from \gets \emptyset$\;
$closed \gets \emptyset$\;

\While{\textit{openSet} not empty}{
    Pop $(f, g_u, u)$ with smallest $f$ from \textit{openSet}\;
    \If{$u = t$}{\textbf{break}}
    \If{$u \in closed$}{\textbf{continue}}
    Add $u$ to $closed$\;

    \ForEach{$(di, dj, cost) \in N_8$}{
        $v \gets (u.i + di,\ u.j + dj)$\;
        \If{$v$ out of bounds or $G[v] = 0$}{\textbf{continue}}
        $g_v \gets g_u + cost$\;
        \If{$g_v < g[v]$ (default $\infty$)}{
            $g[v] \gets g_v$\;
            $f_v \gets g_v + h(v, t)$\;
            Push $(f_v, g_v, v)$ into \textit{openSet}\;
            $came\_from[v] \gets u$\;
        }
    }
}

\Return \textsc{ReconstructPath}$(came\_from, s, t)$\;

\vspace{1em}
\textbf{Function} \textsc{ReconstructPath}$(came\_from, s, t)$\;
$path \gets [t]$\;
$node \gets t$\;
\While{$node \neq s$}{
    $node \gets came\_from[node]$\;
    \If{$node$ undefined}{\Return empty list}
    Prepend $node$ to $path$\;
}
\Return $path$\;

\end{algorithm}

\subsection{Path Compression}
\begin{figure}[!ht]
    \centering
    \begin{subfigure}[a]{1\textwidth}
    \centering
    \includegraphics[width=0.7\linewidth]{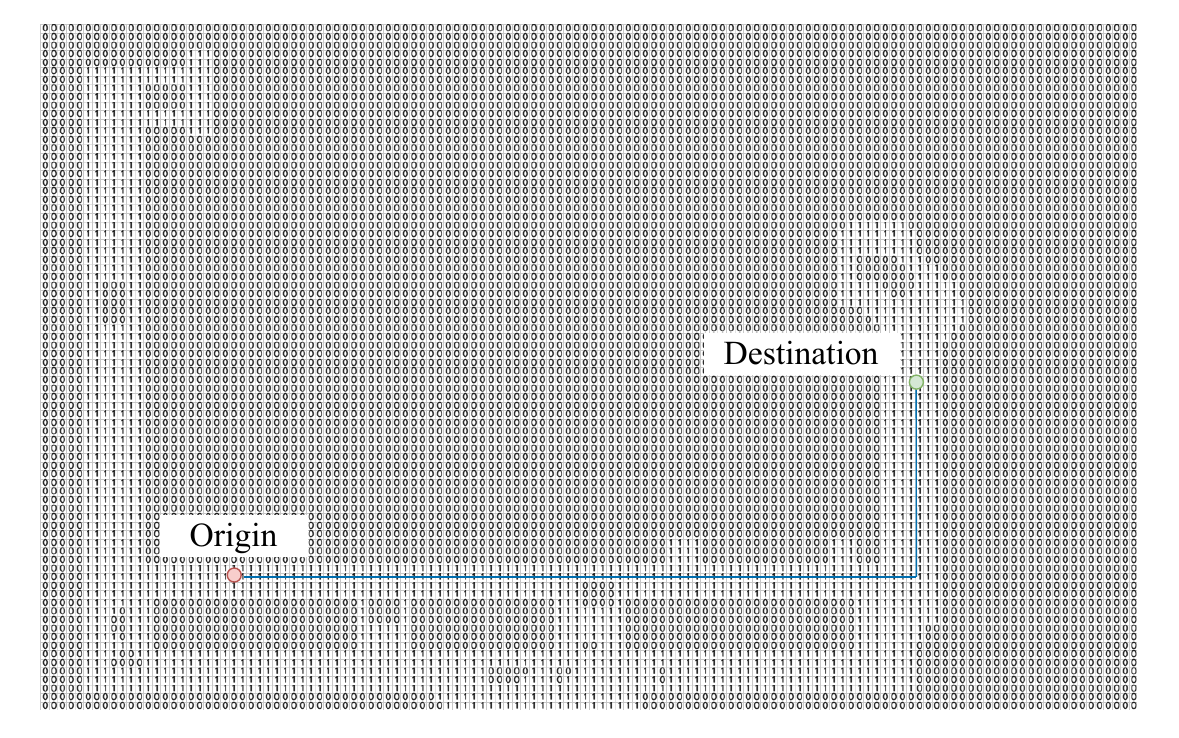}
    \caption{}
    \label{fig:normal_a_star}
    \end{subfigure}
    \begin{subfigure}[b]{1\textwidth}
    \centering
    \includegraphics[width=0.7\linewidth]{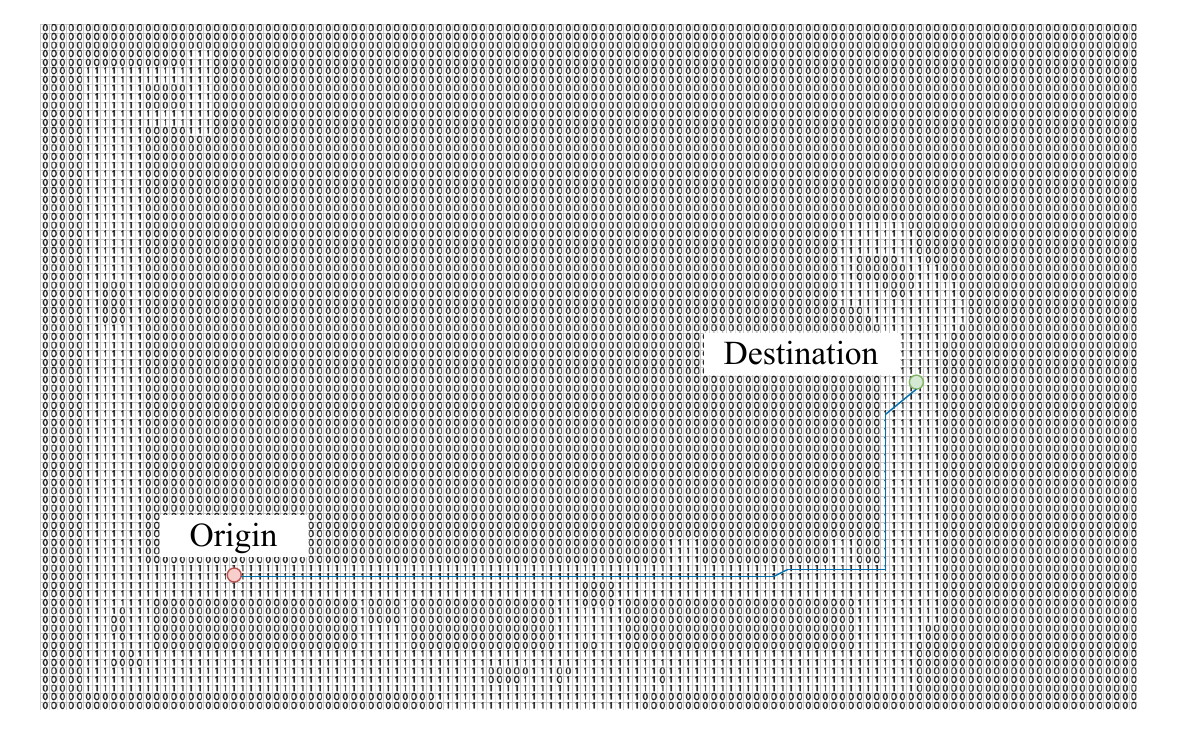}
    \caption{}
    \label{fig:diagonal_a_star}
    \end{subfigure}
    \caption{(a) A* Search with only orthogonal moves. (b) A* Search with both orthogonal and diagonal moves.}
\end{figure}

The raw output of our A* search algorithm is a long sequence of grid-cell coordinates:

  \begin{equation*}
  \{u_0, u_1, ..., u_L\},
  \end{equation*}

where each $u_k = (i_k, j_k)$ denotes the $k$th cell along the given optimal path. To transform this verbose list into concise movement commands, we apply a three-stage compression pipeline:
\begin{itemize}
    \item Stage 1 - Vectorization: We first computed the displacement between each successive pair of cells:
    
      \begin{equation*}
      \Delta u_k = u_{k+1} - u_k,
      \end{equation*}

    which yields one of the eight possible unit vectors \{(-1,0), (1,0), (0,-1), (0,1), (-1, -1), (-1,1), (1,-1), (1,1)\}. Each vector is mapped to a compass code: 'N', 'S', 'E', 'W', 'NE', 'NW', 'SE', or 'SW'.

    \item Stage 2 - Run-Length Encoding: The resulting sequence of direction code often contains long stretches of the same move (e.g., nine steps east). We performed a single linear scan over the direction list and merged consecutive identical codes into tuples $(\text{dir}, n\}$, where $n$ is the count of repeated steps. For example:
    
      \begin{equation*}
      [E, E, E, E, E, S, S] \rightarrow [(E,4), (S,2)]
      \end{equation*}

    This step cost can be mapped with the grid size to convert to step-counts or distance-to-walk in the real-world scenario.

    \item Stage 3 - Diagonal Collapse: Zig-zag walking patterns are unrealistic and need to be avoided. To address that, we scanned the run-length list for adjacent single-step pairs such as $(S,1)$ followed by $(E,1)$. Whenever found, we replaced those with a single diagonal entry such as $(SE, 1)$. We then performed a second pass of run-length encoding to merge any resulting consecutive diagonal moves (e.g., combining two $(SE,1)$ into $(SE,2)$). A sample representation of the diagonally collapsed path and comparison with only the orthogonal path are shown in Figure \ref{fig:normal_a_star} and \ref{fig:diagonal_a_star}.
\end{itemize}

This compression runs in $O(L)$ time, where $L$ is the length of the original map. The total compression overhead is negligible.

Final instruction list after compression might appear as:

  \begin{equation*}
  [('SE', 5), ('N', 3), ('E', 2)]
  \end{equation*}

which we can render in text as:
\begin{quote}
    {Go SE 5 steps}\\
    {Go N 3 steps}\\
    {G E 2 steps}
\end{quote}

\subsection{SLM-based Instruction Generation}
To transform terse commands into natural, numbered directions, we integrate an SLM. The SLM is instruction-tuned with a small dataset. On top of that, a system prompt is provided to give the model some context. This will help the model to generate simple, accurate, and human-understandable instructions and remove unnecessary hallucinations. We assemble a complete prompt incorporating the terse outputs from the A* Search. Hence, the SLM prompt has two parts: (i) System Prompt: To reduce model hallucination and eliminate common mistakes. This needs to be model-specific and requires an iterative approach to be refined, eliminating common mistakes of the particular SLM; and (ii) Terse Output from A*: Point-to-point navigation output received from the A*, along with cost for the navigation. To improve the performance of the model, we implement the following three approaches: (i) Tokenization: Tokenizing the prompt into sub-word units; (ii) Model Invocation: Invoking the model with tuning different parameters, such as the upper bound of generated tokens, randomness, and sampling filter values; and (iii) Post-processing: Ensuring proper output format with a simple regex check.

\section{Evaluation}
We designed an experimental framework to evaluate each of the five pipeline stages. In the following sections, we describe the setup and measurement procedures for map preprocessing and grid generation, graph encoding, 8‑way A* search, path compression, and language‑model‑based instruction synthesis. The average runtime of our approach and navigation instruction accuracies are also reported in this section.  We conducted experiments on a desktop computer. Table \ref{tab:system_configuration} provides the configuration of the desktop computer. To be noted, all the model inferences and processings are done specifically using the CPU.

\begin{table}[!ht]
	\caption{Experimentation system configuration}
	\label{tab:system_configuration}
	\begin{center}
		\begin{tabular}{l l}
			Component & Specification \\\hline
			CPU & AMD Ryzen 7 5700X 8C16T\\
			GPU & AMD Radeon RX 6600 8GB GDDR6\\
			RAM & 32 GB \\
			Operating System & Microsoft Windows 11 \\\hline
		\end{tabular}
	\end{center}
\end{table}

\subsection{Map Preprocessing and Grid Generation}
\begin{figure}[!ht]
    \centering
    \begin{subfigure}[a]{0.8\textwidth}
    \centering
    \includegraphics[width=0.6\linewidth]{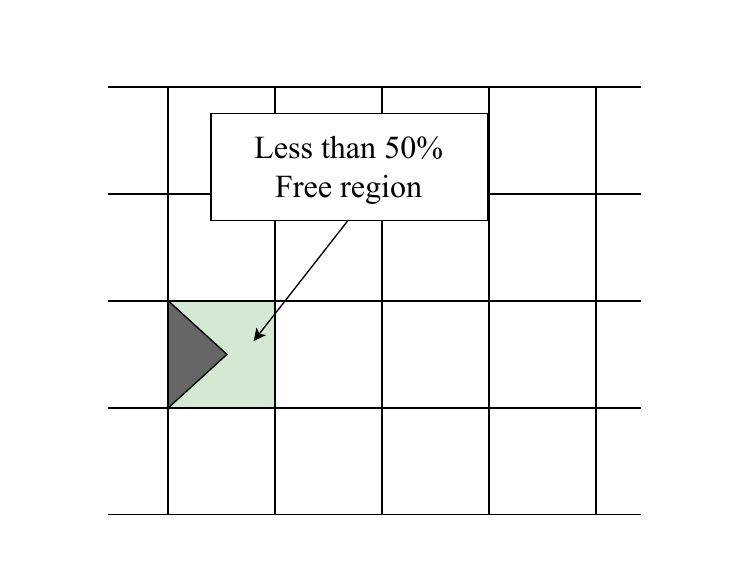}
    \caption{}
    \label{fig:free_region}
    \end{subfigure}
    \begin{subfigure}[b]{0.8\textwidth}
    \centering
    \includegraphics[width=0.6\linewidth]{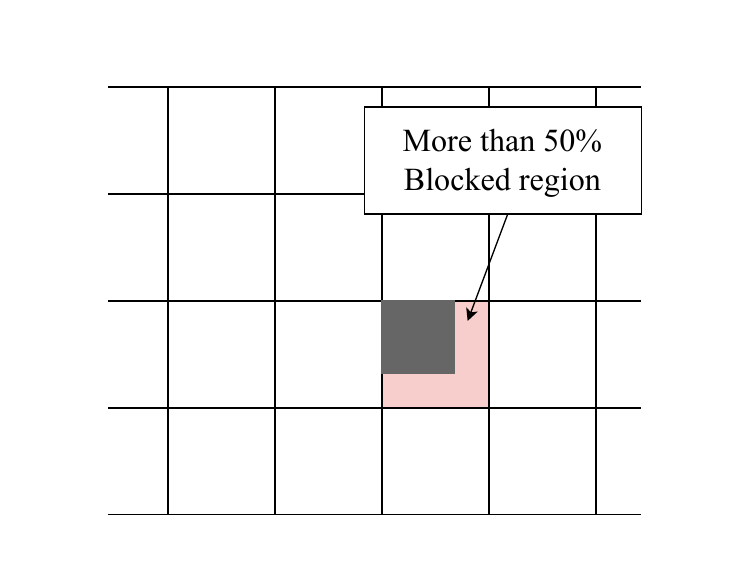}
    \caption{}
    \label{fig:blocked_region}
    \end{subfigure}
    \caption{(a) The square has less than 50\% obstacles from the map, hence the square is decided to be free/walkable. (b) The square has more than 50\% obstacles from the map, hence the square is decided to be blocked/non-walkable.}
\end{figure}

For our experimentation, we considered publicly available existing indoor maps of four different places: (i) Birmingham-Shuttlesworth International Airport, Alabama, USA. (ii) Milan Bergamo Airport, Bergamo, Italy; (iii) Bologna Guglielmo Marconi Airport, Bologna, Italy; and (iv) Orio Center, Bergamo, Italy. High-quality map images were collected for all these indoor locations. For grid generation, we utilized a Python script to load the map images as grid overlays of specific dimensions for each map. The grid dimensions were chosen based on the aspect ratios of the collected map images. Table \ref{tab:grid_dimensions} shows grid dimensions for each map images used in experimentation. An interactive grid layer was created over the map overlays to mark the walkable paths and blocked regions on the grid. We load an 8‑bit grayscale image and partition it into the same $H \times W$ cell grid used during interactive editing. For each cell, we examine the corresponding pixel block in the map overlay: if more than half of its pixels are blocked by any obstacles in the actual map, we mark the cell blocked (0); otherwise, we mark it free (1). This majority‑voting strategy handles noisy edges or slight misalignments in the hand‑drawn mask, and guarantees that only cells designated as walkable remain passable. An example case of this is presented in Figure \ref{fig:free_region} and \ref{fig:blocked_region}. 

\begin{table}[!ht]
	\caption{Grid dimensions of different map images in experimentation}
	\label{tab:grid_dimensions}
	\begin{center}
		\begin{tabular}{l l l}
			Map & Image Aspect Ratio & Grid Dimension $(H \times W)$ \\\hline
			Birmingham & $1292:903$ & $90 \times 130$ \\
			Bergamo & $1074:317$ & $30 \times 107$ \\
			Bologna & $1876:1165$ & $120 \times 190$ \\
			Orio Center (Ground Floor) & $1281:809$ & $80 \times 130$ \\
                Orio Center (First Floor) & $1169:947$ & $80 \times 100$ \\\hline
		\end{tabular}
	\end{center}
\end{table}

\subsection{Graph Encoding}
\begin{figure}[!ht]
  \centering
  \includegraphics[width=.65\linewidth]{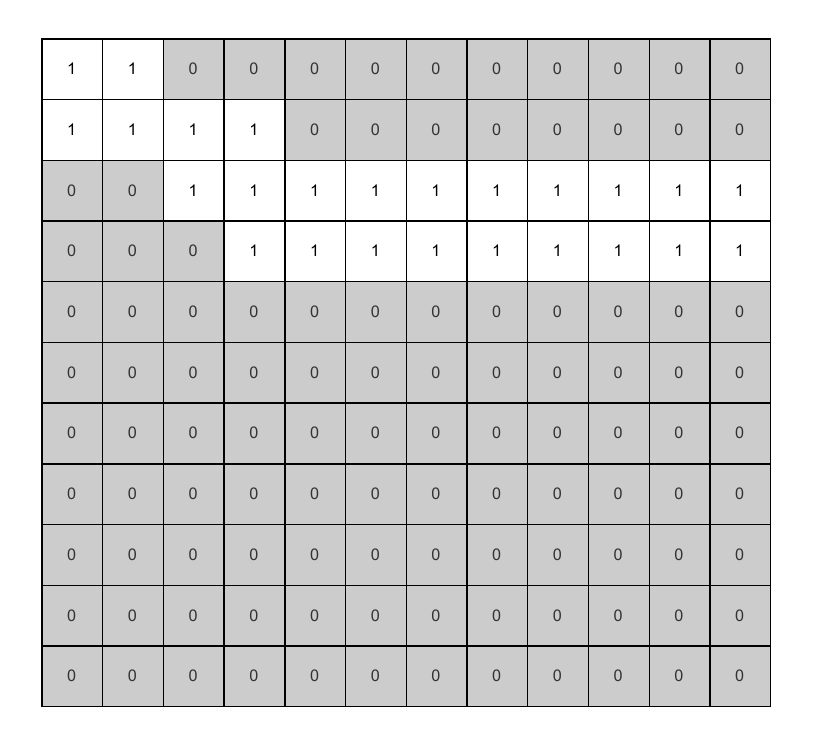}
  \caption{A sample $11 \times 12$ occupancy matrix $G_{11, 12}$. Here, $1$ denotes free/walkable regions, and $0$ denotes blocked/non-walkable paths.}
  \label{fig:occupancy_matrix}
\end{figure}

After converting each map image into a regular grid mask, we construct a Boolean occupancy matrix in Python: free (walkable) cells are encoded as 1 and blocked (non‑walkable) cells as 0. Behind the scenes, we leveraged NumPy’s slicing and vectorized comparisons to build this matrix. A sample $ 11 \times 12$ occupancy matrix is shown in Figure \ref{fig:occupancy_matrix}.

Along with encoding the walkable and non-walkable regions, the coordinates of different locations and landmarks (shops, terminals, etc.) were also extracted. These will be used as the input for the source and the destination for the A* search. For instance, for the map of the Birmingham Airport, the coordinates for the Southwest gate are $(x=53, y=16)$ on a $90 \times 130$ occupancy matrix.

\subsection{Running A* Search with Path Compression}
\begin{figure}[!ht]
  \centering
  \includegraphics[width=0.9\textwidth]{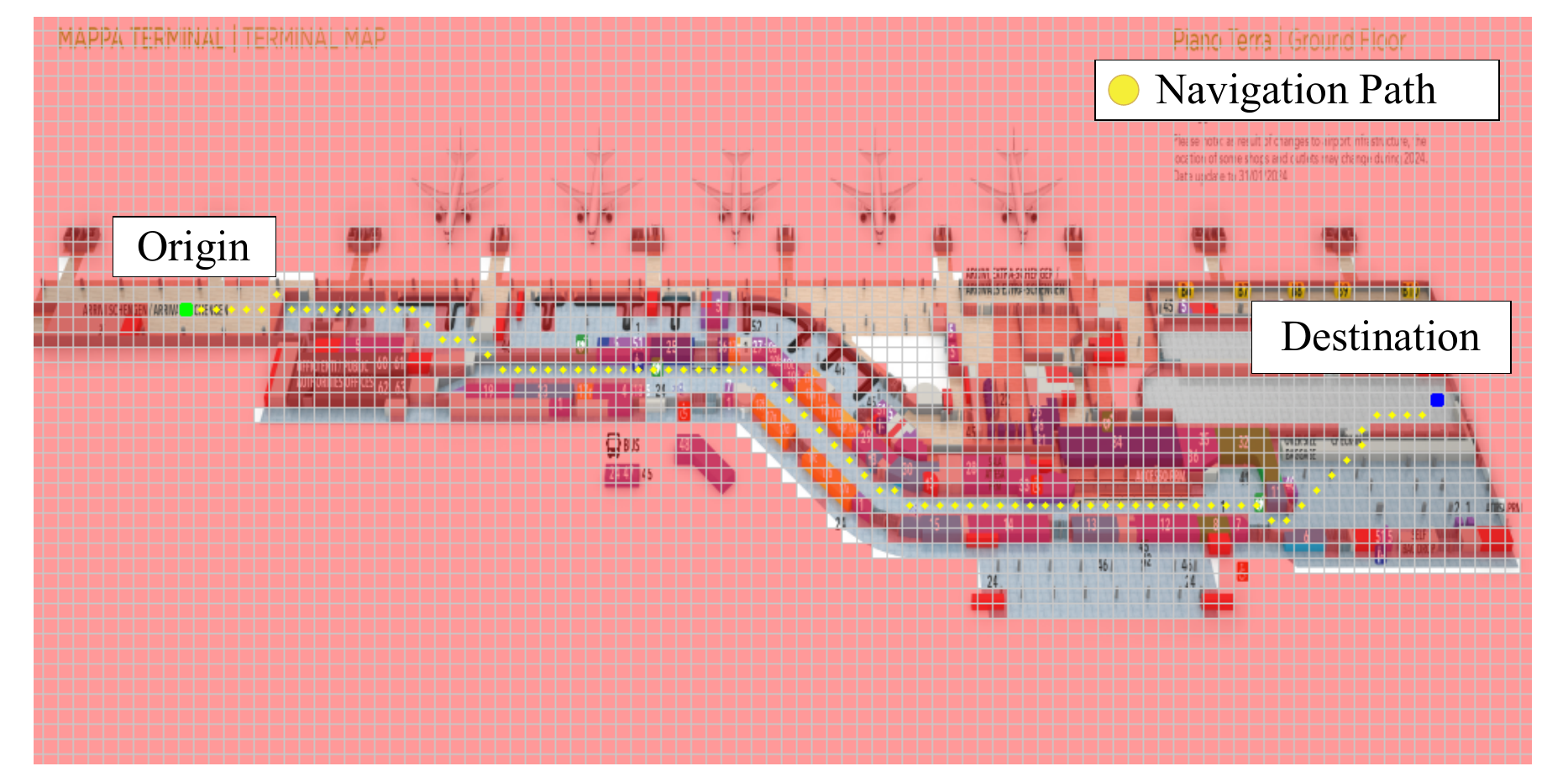}
  \caption{A* Pathfinding on Bologna Guglielmo Marconi Airport Map.}
  \label{fig:sample_a_star}
\end{figure}

The A* Search algorithm has the following input and output - (i) Input: coordinates of the source and the destination; and (ii) Output: navigation instructions with required cost. With the Boolean grid in memory, we invoked the A* routine to compute the shortest paths. Our implementation treats each free cell as a graph node and dynamically enumerates up to eight neighbor offsets: north, south, east, west $(cost=10)$ and the four diagonals $(cost = \sqrt{2})$. We use the Chebyshev distance (i.e., $max\{|\Delta x|, |\Delta y|\}$ as our admissible heuristic. Figure \ref{fig:sample_a_star} shows an example of A* pathfinding result on the Bologna Guglielmo Marconi Airport Map.

We have run a total of 80 test cases on all four collected maps. For each test case, we supply the start and goal cell indices. On a pure CPU run, A* completes under 5 milliseconds on average in all the maps in different experiments, as presented in \ref{tab:a_star_time}. We ran a total of 20 different experiments with different input combinations on each of the 4 maps.

\begin{table}[!ht]
	\caption{Average runtime (in milliseconds) to perform A* search on different maps}
	\label{tab:a_star_time}
	\begin{center}
		\begin{tabular}{l l}
			Map & Average Time (A*) \\\hline
			Birmingham & $0.77\,\text{ms}$ \\
			Bergamo & $1.18\,\text{ms}$ \\
			Bologna & $4.14\,\text{ms}$ \\
			Orio Center (Ground Floor) & $4.23\,\text{ms}$ \\
                Orio Center (First Floor) & $4.09\,\text{ms}$\\
                Orio Center (Multi-floor) & $8.80\,\text{ms}$\\\hline
		\end{tabular}
	\end{center}
\end{table}

The relatively simple layouts of the Birmingham and Bergamo maps translate into shorter average search times, whereas the larger, more intricate floor plans of Bologna and Orio Center naturally require more processing. Because A* is a fully deterministic graph search, it produces the exact same optimal path on every run, eliminating the variability of solely relying on a language model for wayfinding, and it computes these routes in just a few milliseconds, making it ideal for real‑time indoor navigation.

\subsection{SLM based Instruction Synthesis}
Our final stage converted the terse run-length commands into numbered walking instructions. Terse navigation output from the A* was provided to the SLM as input.

For our experimentation, we used TinyLlama-1.1B, a small local language model with approximately 1.1 billion parameters, optimized for resource-constrained environments. This model supports multi-turn instruction-following and conversational dialogue generation, making it well-suited for natural language generation tasks \cite{zhang2024tinyllama}. The variant used in our experimentation has been fine-tuned using parameter-efficient fine-tuning (PEFT) techniques, specifically Low-Rank Adaptation (LoRA). This allowed us to specialize the model for this particular task while keeping memory and compute requirements reasonable.  We trained the model on 1000 sample data points of converting terse commands into conversational output. All the inferences in our experimentation were performed using the CPU. Table \ref{tab:total_time} presents the average time in seconds required to run each query (A* Search + SLM) for different maps. It is evident that the whole process of A* pathfinding and SLM output generation takes 14-15 seconds on average.

A carefully designed system prompt played a crucial role in controlling the model's behavior. As the model does not have access to any external knowledge or context beyond the prompt, we iteratively refined it to address common generation errors and ambiguities. System prompt used for our experiments:
\begin{quote}
\textit{You are a precise navigation assistant. 
    Convert provided terse commands into a numbered walking guide.
    Use “Start by walking…” for step 1,
    “Then walk…” for steps 2…(n-1),
    "Take the escalator from Floor n to n+1",
    and “Finally walk…” for the last step.
    Output one numbered line per command.}

    \textit{Here is an example case given:
    Terse commands:
    Go East 3 steps
    Take the escalator from Floor 0 to 1
    Go North 1 step}

   \textit{ For the example terse commands, the output is:
    1. Start by walking east for 3 steps.
    2. Take the escalator from Floor 0 to 1.
    2. Finally, walk north for 1 step, and you will reach your destination.}

    \textit{Terse commands are given as follows: you have to convert them into a numbered list of directions as provided in the prior example. Keep the step number as given; do not modify it. Do not change the order of the steps, and do not add any additional steps. The output should be numbered lines starting with a number followed by a period and a space.}
    
\end{quote}

\begin{table}[!ht]
	\caption{Average runtime (in seconds) to perform A* search with TinyLlama on different maps}
	\label{tab:total_time}
	\begin{center}
		\begin{tabular}{l l}
			Map & Average Time (A* + TinyLlama) \\\hline
			Birmingham & $14.31\,\text{s}$ \\
			Bergamo & $15.82\,\text{s}$ \\
			Bologna & $16.43\,\text{s}$ \\
			Orio Center (Ground Floor) & $16.96\,\text{s}$ \\
                Orio Center (Multi-floor) & $21.09\,\text{s}$ \\
            Average  & $16.92\,\text{s}$ \\ \hline
		\end{tabular}
	\end{center}
\end{table}

Although we utilized TinyLlama in our experiments for its lightweight and deployable nature, the process is entirely text-based and model-agnostic. This can be easily replicated with any instruction-tuned SLM capable of local inference using very limited computational power.

\subsection{Evaluation Outcomes}
To assess the effectiveness of our proposed indoor navigation pipeline, we conducted a comparative evaluation against a closely related method introduced by Coffrini et al. \cite{coffrini2025methodllmenabledindoornavigation}, which serves as our baseline. Their framework represents one of the most comparable approaches in terms of objectives and application context, making it an appropriate reference point for benchmarking performance.

In the subsequent sections, we first present an overview of Coffrini et al.'s methodology, highlighting key architectural components and mechanisms. We then outline our own system’s design and implementation features, emphasizing the fundamental differences in how pathfinding and instruction generation are handled. Finally, we provide a detailed comparison of experimental results, examining metrics such as navigation accuracy, processing time, and scalability across diverse indoor environments.

\subsubsection{Baseline Method}
Coffrini et al. \cite{coffrini2025methodllmenabledindoornavigation} proposed an approach for generating indoor navigation instructions using LLM, specifically ChatGPT. In their implementation, the raw floor plan images of indoor circuits were preprocessed to remove any unnecessary information (texts, legends, etc.). They processed this image further by removing any extraneous visual elements as well, converting it to a simple graph-like representation of the connected components. This image, along with an iteratively refined system prompt and a user prompt specifying the source and destination, was provided as input to the LLM to generate human-readable navigation instructions.

While the method demonstrated a single-instruction-level accuracy of around 86\%, it suffers from significant computational overhead. Specifically, the response time for generating a complete set of navigation instructions remains high, with an average processing time of approximately 4-5 minutes per query, mainly due to the image processing and reasoning capabilities required by the LLM \cite{coffrini2025methodllmenabledindoornavigation}.

\subsubsection{Our Approach:}
We conducted a total of 80 route-finding experiments, evenly distributed across the four test sites. Because our A* search is fully deterministic, it produced the exact same optimal path, and hence the exact same navigation instructions on every trial. In contrast, the purely LLM‐based approach of Coffrini et al. often fails to match the intended floorplan geometry due to the model's probabilistic nature, leading to incorrect or incomplete guidance. A comparison is shown in Table \ref{tab:accuracy}.

\begin{table}[!ht]
	\caption{Comparison of successful route-generation rate}
	\label{tab:accuracy}
	\begin{center}
		\begin{tabular}{l l l}
			Map & Coffrini et al. Accuracy \cite{coffrini2025methodllmenabledindoornavigation} & Our Accuracy \\\hline
			Bergamo & 76.36\% & 100\% \\
			Bologna & 82.09\% & 100\% \\
			Orio Center (Ground Floor) & 62.50\% & 100\% \\
                Orio Center (Multi-floor) & N/A & 100\%\\\hline
		\end{tabular}
	\end{center}
\end{table}

\begin{figure}[!ht]
  \centering
  \includegraphics[width=\linewidth]{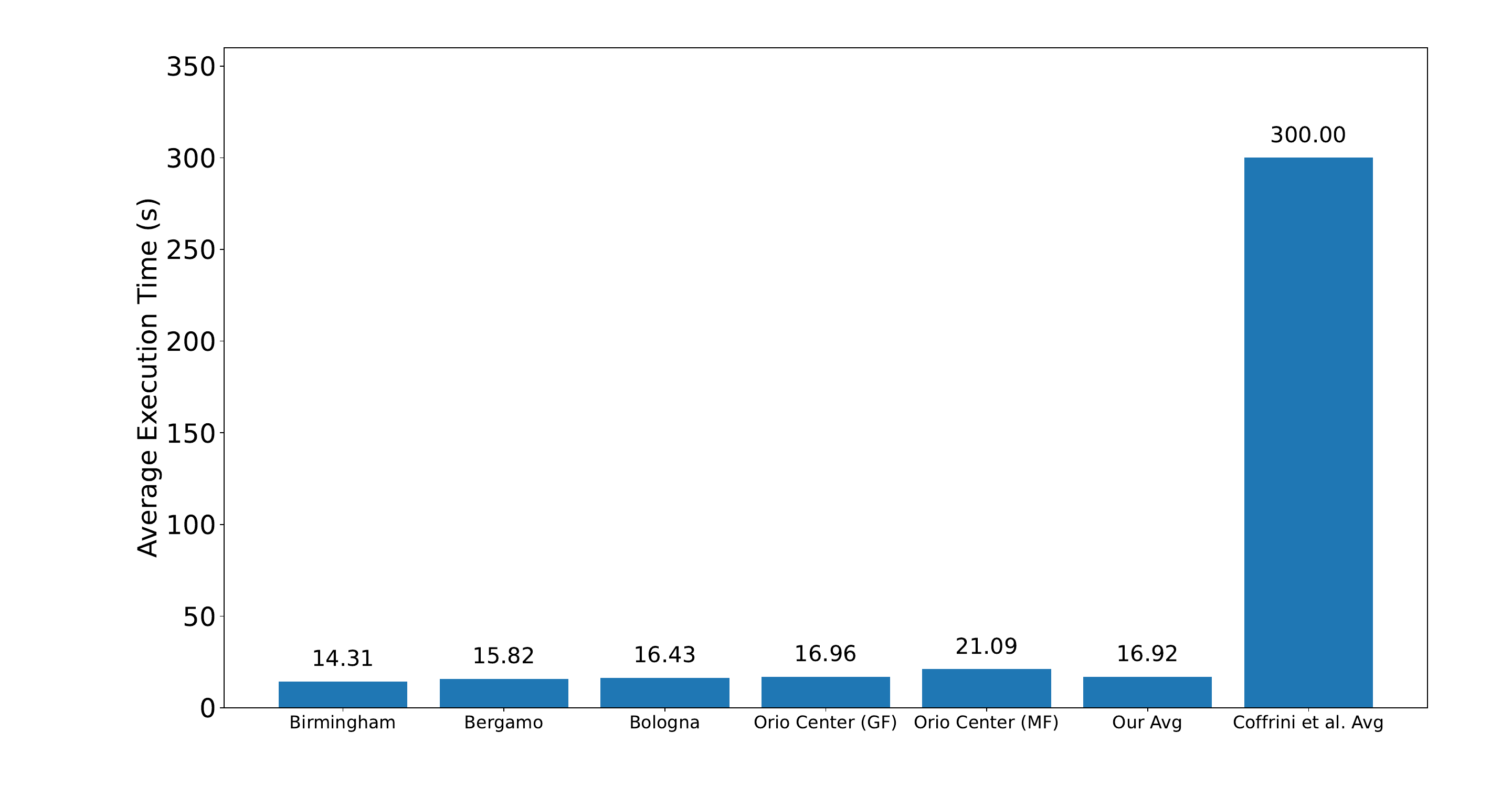}
  \caption{Average execution time (in seconds) comparison.}
  \label{fig:exec_time}
\end{figure}

In their evaluation, Coffrini et al. \cite{coffrini2025methodllmenabledindoornavigation} reported an accuracy of only 76\% on the Bergamo floor plan, with performance declining to 62\% on the larger and complex Orio Center layout. In contrast, our approach integrates A* pathfinding with an SLM, leveraging the A* algorithm for accurate path computation while reserving the machine learning model solely for the generation of natural language instructions. This modular design decouples the image processing and path planning from the language generation, ensuring the reasoning over spatial structures is handled algorithmically rather than probabilistically. As a result, our system achieved 100\% accuracy across all experimented maps. Moreover, the reliance on lightweight local inference substantially reduced computational overhead. The average time required by A* to compute the optimal path is under 5 milliseconds, as shown in Table \ref{tab:a_star_time}, and the total average time per iteration, including both pathfinding and SLM text generation, remains under 20 seconds (Table \ref{tab:total_time}). This demonstrates practical usability in real-world scenarios. A comparison of processing times is shown in Figure \ref{fig:exec_time}. 

Taken together, these results confirm that by offloading deterministic path planning to A* and limiting the machine learning model’s role to lightweight text formatting, we can deliver both 100\% route accuracy and true real‑time performance, something that purely LLM‑centric pipelines cannot match on resource‑constrained devices.

\section{Conclusion}
This paper introduces a new A* enabled SLM indoor navigation approach for personal navigation using handheld devices. Our objective is to provide real-time, accurate point-to-point navigation from grid maps converted from floor plans for an indoor environment without requiring any specialized hardware infrastructure. Merely relying on language models to process the map image and provide navigation instructions can be time-consuming and resource-intensive, and the probabilistic output of a machine learning model is more prone to errors. 

In this study, by converting standard architectural floor-plan images into lightweight binary occupancy grids, we eliminate the need for heavy image processing inside the LLM. Our 8-way A* implementation then computes optimal routes in a few milliseconds, ensuring real-time response on commodity hardware. A language model is only utilized for converting the terse texts into conversational, human-understandable output. Moreover, an SLM can be trained further to provide more customized output for different contexts. The text outputs can be easily converted to speech instructions, providing directional guidance to people with visual impairments.

Future work should involve finding optimized navigation instructions dynamically. In practical settings, corridors and entrances may become temporarily blocked by crowds, maintenance works, or barriers. Automated grid map creation with sensor outputs can reflect such updates into the occupancy grid.

We used the generative AI tool `ChatGPT' to help rephrase parts of our own writing to improve clarity and for editorial purposes.

\section{Authors Contribution}
\textbf{Md. Wasiul Haque, Sagar Dasgupta:} conceptualization, methodology, coding, data collection, data analysis, and writing – original draft; \textbf{Mizanur Rahman:} conceptualization, methodology, writing – original draft, review and editing, and funding acquisition.

\section*{acknowledgements}
This work is based upon the work supported by the National Science Foundation (NSF) (Award \# 2340456). Any opinions, findings, conclusions, and recommendations expressed in this material are those of the author(s) and do not necessarily reflect the views of NSF, and the U.S. Government assumes no liability for the contents or use thereof.

% the apacite bibliography style matches the ION bibliography style guidelines.
\bibliographystyle{apalike}
\bibliography{citations.bib}

\end{document}